\documentclass[12pt]{article}

\usepackage{sbc-template}
\usepackage{graphicx,url}
\usepackage[utf8]{inputenc}
\usepackage{multirow}
\usepackage{booktabs}
\usepackage{amsmath}
\usepackage{xcolor}
\usepackage{url}
\usepackage{hyperref}

\title{A Coreset Selection Framework with \\ Ensemble Aggregation for Image Classification}

\author{Pedro Rocha Dantas\inst{1}, Lucas Pascotti Valem\inst{1}}

\address{
Institute of Mathematics and Computer Science (ICMC) \\
University of São Paulo (USP) \\
São Carlos -- SP -- Brazil
\email{pedrord@usp.br, lucas@icmc.usp.br}
}

\begin{document} 

\maketitle

\begin{center}
\vspace{-2mm}
\fbox{\parbox{0.92\columnwidth}{\centering\small
\textbf{Preprint.} Accepted at the Workshop de Trabalhos de Alunos de Graduação (WTAG) of the Brazilian Symposium on Databases (SBBD~2026).}}
\vspace{2mm}
\end{center}

\begin{abstract}
The rapid growth of image data has produced large-scale datasets, raising concerns about the time and memory costs of model training. Selecting representative training subsets, however, remains challenging: individual sample contributions are unclear, and model behavior varies across datasets and runs. We address these challenges with a framework that combines coreset selection with an ensemble aggregation over multiple runs. For coreset selection, we propose SCOre-Stratified Selection (SCOSS), which partitions the training data into intervals based on a chosen score and samples from each interval. The ensemble combines predictions from multiple runs, each performed on an independently sampled training subset. As baselines, we use moderate and random selection, each in original and class-balanced versions. We assess the framework with Simple Graph Convolution (SGC) and Support Vector Machine (SVM) classifiers under different sampling ratios. Experiments show that SCOSS is competitive with baselines, often the best choice for SGC, and enables favorable trade-offs between accuracy and efficiency. On the fine-grained dataset, SGC with SCOSS outperforms SVMs when using fewer labeled samples. The code and supplementary materials are publicly available at \href{http://scoss.lucasvalem.com}{\texttt{scoss.lucasvalem.com}}.
\end{abstract}

\section{Introduction}

The growing availability of large-scale datasets has driven progress in image classification, but has also raised concerns about the computational cost of model training in terms of memory and processing time~\cite{moser2026coresetselectioncoresetselection}. This concern is even more pressing in transductive models such as Graph Convolutional Networks (GCNs)~\cite{surveyGNN2024IEEE}, where the entire graph structure participates in training and inference. Therefore, reducing the amount of training data while preserving classification performance has become a central challenge in the field~\cite{deepcore}.

A common strategy to address this problem is coreset selection, which aims to identify representative subsets of the original dataset to reduce training cost~\cite{feldman2020_introduction, deepcore}. Despite its potential, selecting informative samples remains difficult, as the contribution of each instance to model performance is not always clear, especially in graph-structured data where samples are interdependent~\cite{paperGCN-SGC2019}. Moreover, model behavior may vary across datasets and training runs, making the selection process inherently unstable. These challenges motivate the search for approaches capable of producing representative subsets while improving robustness and consistency across executions~\cite{moser2026coresetselectioncoresetselection, Sagi2018_ensemble}.

Several approaches have been proposed for coreset selection in classification tasks \cite{feldman2020_introduction}. Random sampling and Moderate Coreset \cite{xia2023_moderate} serve as common baselines: the former selects instances uniformly at random, while the latter focuses on samples close to the score median. More structured methods include one inspired by stratified sampling that improves data coverage \cite{Haizhong2023_Coverage}, and another that allocates the budget per class based on within-class difficulty \cite{Elisa2025_Class}. The method proposed in this work combines both ideas, applying stratified sampling over the full score distribution with class-balanced budget allocation.

In this work, we propose a framework that combines coreset selection with an ensemble strategy to improve robustness and predictive performance in image classification. We introduce SCOre-Stratified Selection (SCOSS), a method that partitions the training data into score-based intervals and samples instances from each interval. Predictions from multiple runs are then aggregated through an ensemble mechanism to reduce variability. We evaluate the approach using Simple Graph Convolution (SGC) \cite{paperGCN-SGC2019} and Support Vector Machine (SVM) under different sampling ratios, comparing against random sampling and Moderate Coreset baselines, including their class-balanced variants. Results indicate that SCOSS is competitive with existing methods and frequently achieves the best performance for SGC, particularly on a fine-grained dataset with many classes and few samples per class.

\section{Methodology}

This section presents the proposed framework for coreset selection, the SCOre-Stratified Selection (SCOSS) algorithm, and the baseline methods used for comparison.

\subsection{Framework Overview}

\begin{figure}[ht]
    \vspace*{-1mm}
    \centering
    \includegraphics[width=\linewidth]{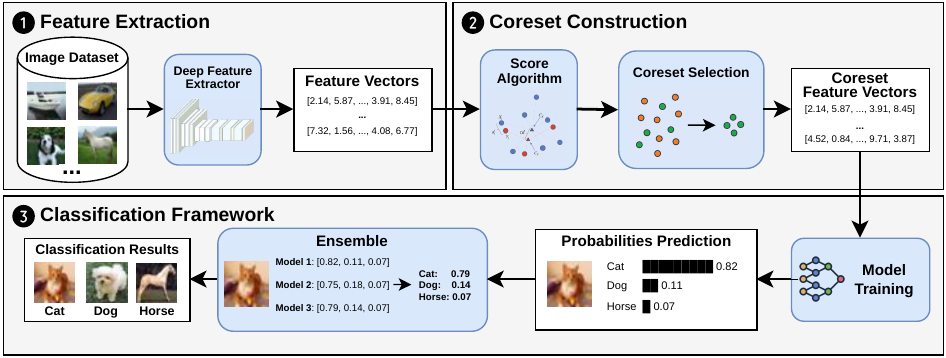}
    \caption{Overview of the proposed framework for coreset selection.}
    \label{fig:workflow1}
    \vspace*{-3mm}
\end{figure}

Figure~\ref{fig:workflow1} illustrates the workflow of the proposed framework, composed of three stages: feature extraction, coreset construction, and classification with ensemble aggregation.

Initially, each image is mapped to a fixed-dimensional feature vector using a pretrained deep neural network. These representations capture semantic information and serve as the basis for both coreset construction and classification in the subsequent stages.

The second stage is the coreset construction process, the central component of the proposed approach. Given the extracted feature vectors, a score is computed for each image in the training set according to a predefined scoring criterion. In this work, the Euclidean distance to the class centroid, defined as the mean feature vector of all samples in that class, is adopted as the scoring criterion. These scores reflect how representative each image is within its class and are computed once over the full training set prior to any selection and are then passed to the coreset selection method to select a representative subset of the training data. Both SCOSS and the baseline methods operate on these scores and are described in the following subsections.

Finally, the selected coreset is used to train classification models such as SGC and SVM. Since coreset selection involves random sampling, predictions may vary across runs. To reduce this variability, the framework optionally incorporates an ensemble mechanism that performs multiple independent runs, each sampling a distinct coreset. Predictions from each run are aggregated through an ensemble strategy, which can take different forms~\cite{Sagi2018_ensemble}. In this work, we adopt simple probability averaging. Both settings (i.e., with and without ensemble) are explored in the experiments, and performance is evaluated in terms of accuracy over the full test set.

\subsection{SCOre-Stratified Selection (SCOSS)}

We propose SCOre-Stratified Selection (SCOSS), a coreset selection strategy designed to preserve the overall score distribution of the training data. Unlike methods that focus on a specific region of the score distribution, SCOSS covers the full range of scores through stratified sampling. Three variants are considered: \emph{(i)} a standard version (SCOSS), operating over the entire training set; \emph{(ii)} a class-balanced version (SCOSS$_B$), performing selection independently for each class; and \emph{(iii)} a graph-based version (SCOSS$_{CC}$), which incorporates structural information from a $k$NN graph to promote diversity and exploits connected components from the graph.

\begin{figure}[ht]
    \centering
    \vspace*{-4mm}
    \includegraphics[
        width=\textwidth,
        trim=0 20 0 20,
        clip
    ]{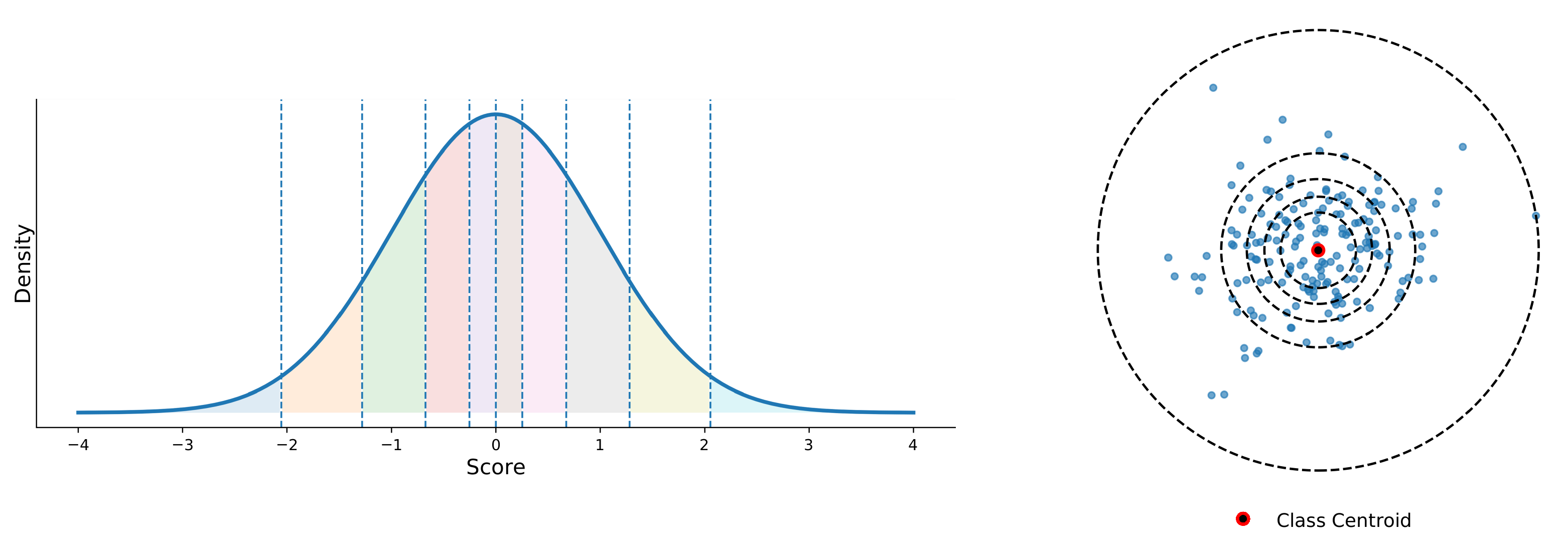}
    \caption{Visual Representation of SCOSS interval division.}
    \label{fig:SCOSS_representation}
    \vspace*{-3mm}
\end{figure}

Given the training set $\mathcal{D}$ with $N$ samples, each associated with a score $s_i$, the standard version of SCOSS operates as follows. The samples are sorted in ascending order by score and partitioned into $M$ equal-sized intervals of $\lfloor N/M \rfloor$ samples each. When scores follow an approximately normal distribution, central intervals span a narrower score range while tail intervals span a wider range, as illustrated in Figure~\ref{fig:SCOSS_representation}. From each interval, $k = \lfloor r \cdot N / M \rfloor$ samples are randomly drawn, where $r \in (0,1]$ is the sampling ratio. Then, the selected samples are combined to form the coreset. If the total is less than $l = \lfloor r \cdot N \rfloor$ due to rounding in the per-interval budget, the remaining $l - M \cdot k$ samples are drawn uniformly at random from the full training set to complete the coreset.

SCOSS$_B$ applies the same stratified sampling strategy independently for each class. The total budget is $l = \lfloor r \cdot N \rfloor$, with a per-class budget of $l_c = \max(1, \lfloor l / C \rfloor)$, where $C$ is the number of classes. The number of intervals for class $c$ is computed proportionally to its score range as $M_c = \max\left(1, \operatorname{round}\left(M \cdot \frac{a_c}{a}\right)\right)$, where $a = \max(s) - \min(s)$ and $a_c = \max(s_c) - \min(s_c)$. For each class $c$, samples are sorted by score, partitioned into $M_c$ equal-sized intervals, and $\lfloor l_c / M_c \rfloor$ samples are randomly drawn from each. If fewer than $l_c$ samples are selected, the remainder are drawn uniformly at random from within the same class. The final coreset combines samples across all classes. If fewer than $l$ samples are obtained, the remaining $l - \sum_{c=1}^{C} l_c$ are drawn uniformly at random from the full training set.

SCOSS$_{CC}$ combines class-balanced stratified selection with a graph-based component to promote structural diversity. A $k$NN graph is constructed over the extracted feature vectors using Euclidean distance, and its connected components are identified. The total budget is split by parameter $p \in (0,1)$: a fraction $\lfloor p \cdot l \rfloor$ is selected via SCOSS$_B$, while the remaining $l - \lfloor p \cdot l \rfloor$ samples are drawn from the connected components, considering only samples not already selected. Components are sorted by size, and only those with size greater than or equal to the median are eligible. One sample is initially drawn uniformly at random from each eligible component to ensure structural coverage, and if budget remains, additional samples are drawn from the remaining nodes of those components. The final coreset combines samples from both stages.

\subsection{Baseline Methods}

Four baseline methods are considered for comparison. Random sampling selects $l$ instances uniformly at random, without considering any notion of importance or class distribution. Moderate Coreset~\cite{xia2023_moderate} selects samples whose scores are closest to the median of the score distribution, focusing only on a specific region. Class-balanced variants of both methods are also considered (indicated by subscript $_{B}$), applying the same strategies independently for each class with a per-class budget of $l_c = \max(1, \lfloor l / C \rfloor)$ samples.

\section{Experimental Evaluation}

Feature vectors were extracted using a ResNet-152~\cite{paperRESNET} pretrained on ImageNet, providing 2048-dimensional vectors from the final convolutional stage. Experiments were conducted on CUB-200~\cite{Wah2011_CUB200}, a fine-grained dataset with 200 classes (i.e., species of birds) and 5,994 training images; and CIFAR-10~\cite{krizhevsky2009_CIFAR10}, with 10 classes and 50,000 training images. Coreset selection was evaluated under sampling ratios of 2.5\%, 5\%, 10\%, and 20\% and performance was measured in terms of accuracy over the full test set.

For SGC, a reciprocal $k$NN graph was built over the feature vectors using Euclidean distance, with $k=40$, $\mathrm{lr}=0.01$, and 200 epochs for CUB-200, and $k=30$, $\mathrm{lr}=0.001$, and 300 epochs for CIFAR-10. For SVM, an RBF kernel with $C=10$ was used on both datasets. All SCOSS variants used $M=16$ intervals, and SCOSS$_{CC}$ fixes $p=0.9$ with the $k$NN graph constructed using $k=5$.
All methods were evaluated over 10 independent runs, with results reported as mean accuracy and standard deviation. Ensemble results were obtained by averaging predictions across 10 runs, with mean and standard deviation computed over 5 ensembles. Training time and GPU memory during SGC training were also reported for each sampling ratio on an NVIDIA RTX 3090 (24 GB), and compared against the full training set baseline. The reported costs include both coreset selection and SGC training.

Tables 1 and 2 report the full training set accuracy and the SGC computational cost (i.e., execution time and GPU memory used) per sampling ratio, respectively. Tables 3 and 4 present the classification accuracy for all methods with and without ensemble on CUB-200 and CIFAR-10, respectively. All results are reported as mean accuracy and standard deviation.
The results reveal an important finding: SCOSS$_B$ achieves the best or a competitive average rank for SGC on both datasets, indicating that covering the full score distribution with class-balanced allocation is an effective strategy.

\begin{table}[ht]
    \centering
    \begin{minipage}[b]{0.40\columnwidth}
        \centering
        \caption{Full training set accuracy (\%).}
        \label{tab:baseline}
        \resizebox{\linewidth}{!}{
        \begin{tabular}{lcc}
            \toprule
            Dataset & SGC & SVM \\
            \midrule
            CUB-200 & $63.57 \pm 0.14$ & $69.24 \pm 0.00$ \\
            CIFAR-10 & $89.17 \pm 0.04$ & $99.73 \pm 0.00$ \\
            \bottomrule
        \end{tabular}}
    \end{minipage}
    \begin{minipage}[b]{0.58\columnwidth}
        \centering
        \caption{SGC training time and GPU \\ memory consumption.}
        \label{tab:cost_analysis}
        \resizebox{\linewidth}{!}{
        \setlength{\tabcolsep}{3pt}
        \begin{tabular}{llccccc}
            \toprule
            Dataset & Cost & 100\% & 20\% & 10\% & 5\% & 2.5\% \\
            \midrule
            \multirow{2}{*}{CUB-200} & Time (s) & 3.30 & 1.32 & 1.26 & 1.08 & -- \\
            & Memory (GB) & 3.07 & 2.15 & 1.97 & 1.87 & -- \\
            \midrule
            \multirow{2}{*}{CIFAR-10} & Time (s) & 16.57 & 4.24 & 2.81 & 1.79 & 1.41 \\
            & Memory (GB) & 11.11 & 4.05 & 3.27 & 2.81 & 2.56 \\
            \bottomrule
        \end{tabular}}
    \end{minipage}
    \hfill
\end{table}

\begin{table}[!h]
\centering
\small

\caption{Classification accuracy (\%) on the CUB-200 dataset for SGC and SVM.}
\label{tab:cub200}

\resizebox{\linewidth}{!}{
\begin{tabular}{l|l|ccc|c}

\hline
\multicolumn{6}{c}{\textbf{CUB-200 Dataset - SGC Classifier}} \\ \hline
Ensemble & Method / Ratio ($r$)
& 20\% & 10\% & 5\% & Ave. rank $\downarrow$ \\ \hline
\multirow{7}{*}{\rotatebox{60}{No Ensemble}}
& Random    
& $50.22 \pm 0.84$ & $41.35 \pm 1.22$ & $32.15 \pm 1.31$ & 6.00 \\
& Random$_B$    
& $52.08 \pm 0.54$ & $45.38 \pm 0.78$ & $35.25 \pm 0.70$ & 4.00 \\
& Moderate  
& $48.41 \pm 0.16$ & $40.45 \pm 0.21$ & $31.50 \pm 0.11$ & 7.00 \\
& Moderate$_B$
& $52.70 \pm 0.17$ & $45.87 \pm 0.10$ & $\mathbf{39.12 \pm 0.14}$ & 1.67 \\
& SCOSS ~~~~~(ours)       
& $50.66 \pm 0.82$ & $42.80 \pm 1.01$ & $35.97 \pm 0.49$ & 4.33 \\
& SCOSS$_{B}$ ~~(ours)        
& $\mathbf{53.07 \pm 0.46}$ & $\mathbf{46.67 \pm 0.73}$ & $36.49 \pm 1.34$ & \textbf{1.33} \\
& SCOSS$_{CC}$ (ours)       
& $52.40 \pm 0.65$ & $44.71 \pm 0.72$ & $35.47 \pm 1.24$ & 3.67 \\

\hline

\multirow{7}{*}{\rotatebox{60}{With Ensemble}}
& Random    
& $58.72 \pm 0.32$ & $44.05 \pm 1.00$ & $23.67 \pm 1.00$ & 6.00 \\
& Random$_B$   
& $59.79 \pm 0.19$ & $56.32 \pm 0.38$ & $48.12 \pm 0.28$ & 2.67 \\
& Moderate  
& $48.33 \pm 0.01$ & $40.50 \pm 0.07$ & $31.55 \pm 0.08$ & 6.67 \\
& Moderate$_B$ 
& $52.75 \pm 0.04$ & $45.89 \pm 0.06$ & $39.11 \pm 0.09$ & 5.33 \\
& SCOSS ~~~~~(ours)        
& $58.95 \pm 0.38$ & $54.10 \pm 0.79$ & $47.77 \pm 0.80$ & 4.00 \\
& SCOSS$_B$  ~~(ours)       
& $\mathbf{60.11 \pm 0.34}$ & $\mathbf{57.16 \pm 0.64}$ & $47.90 \pm 0.57$ & \textbf{1.67} \\
& SCOSS$_{CC}$ (ours)      
& $59.89 \pm 0.54$ & $56.37 \pm 0.51$ & $\mathbf{48.37 \pm 0.82}$ & \textbf{1.67} \\
\hline
\multicolumn{6}{c}{\textbf{CUB-200 Dataset - SVM Classifier}} \\
\hline
Ensemble & Method / Ratio ($r$)
& 20\% & 10\% & 5\% & Ave. rank $\downarrow$ \\ \hline

\multirow{7}{*}{\rotatebox{60}{No Ensemble}}
& Random    
& $48.48 \pm 1.07$ & $33.25 \pm 1.03$ & $21.23 \pm 1.21$ & 6.00 \\
& Random$_B$   
& $53.96 \pm 0.43$ & $43.40 \pm 0.67$ & $21.56 \pm 1.15$ & 3.67 \\
& Moderate  
& $46.22 \pm 0.00$ & $30.99 \pm 0.00$ & $19.93 \pm 0.00$ & 7.00 \\
& Moderate$_B$ 
& $54.22 \pm 0.00$ & $43.87 \pm 0.00$ & $\mathbf{27.55 \pm 0.00}$ & 1.67 \\
& SCOSS ~~~~~(ours)    
& $49.03 \pm 0.77$ & $34.98 \pm 0.98$ & $21.79 \pm 1.29$ & 4.67 \\
& SCOSS$_B$ ~~(ours)     
& $\mathbf{54.63 \pm 0.35}$ & $\mathbf{44.35 \pm 0.68}$ & $22.19 \pm 1.10$ & \textbf{1.33} \\
& SCOSS$_{CC}$ (ours)   
& $53.31 \pm 0.58$ & $37.51 \pm 0.86$ & $21.95 \pm 1.50$ & 3.67 \\

\hline
\multirow{7}{*}{\rotatebox{60}{With Ensemble}}
&Random    
& $63.17 \pm 0.54$ & $53.52 \pm 0.79$ & $35.91 \pm 1.23$ & 5.00 \\
&Random$_B$   
& $64.93 \pm 0.28$ & $60.21 \pm 0.42$ & $\mathbf{42.38 \pm 0.91}$ & 1.67 \\
&Moderate  
& $46.22 \pm 0.00$ & $30.99 \pm 0.00$ & $19.93 \pm 0.00$ & 7.00 \\
&Moderate$_B$ 
& $54.22 \pm 0.00$ & $43.87 \pm 0.00$ & $27.55 \pm 0.00$ & 6.00 \\
&SCOSS ~~~~~(ours)        
& $63.43 \pm 0.41$ & $55.56 \pm 0.67$ & $42.24 \pm 1.38$ & 3.67 \\
&SCOSS$_B$ ~~(ours)        
& $\mathbf{65.41 \pm 0.29}$ & $\mathbf{60.55 \pm 0.27}$ & $42.27 \pm 1.16$ & \textbf{1.33} \\
&SCOSS$_{CC}$ (ours)       
& $64.77 \pm 0.34$ & $57.67 \pm 0.61$ & $41.87 \pm 0.50$ & 3.33 \\

\hline
\end{tabular}
}
\end{table}

\begin{table}[!h]
\centering
\small

\caption{Classification accuracy (\%) on the CIFAR-10 dataset for SGC and SVM.}
\label{tab:cifar10_sgc}

\resizebox{\linewidth}{!}{
\begin{tabular}{l|l|cccc|c}

\hline
\multicolumn{7}{c}{\textbf{CIFAR-10 Dataset - SGC Classifier}} \\ \hline

Ensemble & Method / Ratio ($r$)
& 20\% & 10\% & 5\% & 2.5\% &Ave. rank $\downarrow$ \\ \hline

\hline
\multirow{7}{*}{\rotatebox{50}{No Ensemble}}
&Random    
& $87.82 \pm 0.20$ & $87.14 \pm 0.15$ & $86.47 \pm 0.19$ & $85.41 \pm 0.39$ & 4.00 \\
&Random$_B$   
& $87.83 \pm 0.15$ & $87.21 \pm 0.19$ & $86.42 \pm 0.24$ & $85.34 \pm 0.35$ & 4.00 \\
&Moderate  
& $87.47 \pm 0.09$ & $86.77 \pm 0.12$ & $86.12 \pm 0.06$ & $85.18 \pm 0.06$ & 6.00 \\
&Moderate$_B$
& $87.18 \pm 0.08$ & $86.48 \pm 0.08$ & $85.74 \pm 0.06$ & $84.98 \pm 0.10$ & 7.00 \\
&SCOSS ~~~~~(ours)        
& $87.66 \pm 0.19$ & $\mathbf{87.30 \pm 0.19}$ & $86.49 \pm 0.15$ & $85.43 \pm 0.28$ & 3.00 \\
&SCOSS$_B$ ~~(ours)       
& $\mathbf{87.88 \pm 0.17}$ & $87.26 \pm 0.25$ & $\mathbf{86.53 \pm 0.16}$ & $\mathbf{85.47 \pm 0.38}$ & \textbf{1.38} \\
&SCOSS$_{CC}$ (ours)       
& $87.75 \pm 0.17$ & $87.22 \pm 0.21$ & $\mathbf{86.53 \pm 0.23}$ & $85.46 \pm 0.16$ & 2.63 \\

\hline
\multirow{7}{*}{\rotatebox{60}{With Ensemble}}
&Random    
& $88.71 \pm 0.11$ & $88.21 \pm 0.04$ & $87.74 \pm 0.27$ & $87.26 \pm 0.07$ & 3.50 \\
&Random$_B$
& $88.73 \pm 0.09$ & $88.17 \pm 0.13$ & $87.79 \pm 0.14$ & $87.16 \pm 0.18$ & 3.75 \\
&Moderate  
& $87.56 \pm 0.02$ & $86.85 \pm 0.01$ & $86.06 \pm 0.03$ & $85.27 \pm 0.02$ & 6.00 \\
&Moderate$_B$ 
& $87.17 \pm 0.03$ & $86.52 \pm 0.01$ & $85.82 \pm 0.01$ & $84.98 \pm 0.01$ & 7.00 \\
&SCOSS ~~~~~(ours)         
& $88.69 \pm 0.03$ & $\mathbf{88.29 \pm 0.06}$ & $87.74 \pm 0.01$ & $87.24 \pm 0.02$ & 3.38 \\
&SCOSS$_B$ ~~(ours)     
& $\mathbf{88.77 \pm 0.02}$ & $88.25 \pm 0.06$ & $\mathbf{87.80 \pm 0.13}$ & $87.28 \pm 0.21$ & \textbf{1.50} \\
&SCOSS$_{CC}$ (ours)     
& $88.74 \pm 0.04$ & $88.21 \pm 0.08$ & $87.65 \pm 0.10$ & $\mathbf{87.31 \pm 0.12}$ & 2.88 \\

\hline
\multicolumn{7}{c}{\textbf{CIFAR-10 Dataset - SVM Classifier}} \\
\hline
Ensemble & Method / Ratio ($r$)
& 20\% & 10\% & 5\% & 2.5\% & Ave. rank $\downarrow$ \\ \hline

\multirow{7}{*}{\rotatebox{50}{No Ensemble}}
&Random    
& $89.74 \pm 0.12$ & $88.62 \pm 0.25$ & $\mathbf{87.47 \pm 0.24}$ & $\mathbf{86.27 \pm 0.28}$ & \textbf{2.00} \\
&Random$_B$   
& $89.70 \pm 0.16$ & $88.63 \pm 0.22$ & $87.42 \pm 0.24$ & $86.00 \pm 0.24$ & 3.25 \\
&Moderate  
& $88.04 \pm 0.00$ & $86.84 \pm 0.00$ & $85.86 \pm 0.00$ & $85.02 \pm 0.00$ & 6.00 \\
&Moderate$_B$ 
& $85.88 \pm 0.00$ & $85.49 \pm 0.00$ & $85.32 \pm 0.00$ & $84.48 \pm 0.00$ & 7.00 \\
&SCOSS ~~~~~(ours)        
& $89.73 \pm 0.12$ & $88.60 \pm 0.27$ & $87.36 \pm 0.24$ & $85.98 \pm 0.35$ & 4.25 \\
&SCOSS$_B$ ~~(ours)       
& $\mathbf{89.76 \pm 0.31}$ & $\mathbf{88.65 \pm 0.22}$ & $87.32 \pm 0.32$ & $86.02 \pm 0.28$ & 2.25 \\
&SCOSS$_{CC}$ (ours)     
& $89.75 \pm 0.18$ & $88.61 \pm 0.23$ & $87.43 \pm 0.19$ & $85.96 \pm 0.27$ & 3.25 \\

\hline
\multirow{7}{*}{\rotatebox{60}{With Ensemble}}
&Random    
& $\mathbf{91.26 \pm 0.07}$ & $90.40 \pm 0.16$ & $89.35 \pm 0.12$ & $88.30 \pm 0.14$ & 3.00 \\
&Random$_B$   
& $91.22 \pm 0.09$ & $\mathbf{90.41 \pm 0.10}$ & $89.35 \pm 0.11$ & $88.39 \pm 0.06$ & \textbf{2.63} \\
&Moderate  
& $88.04 \pm 0.00$ & $86.84 \pm 0.00$ & $85.86 \pm 0.00$ & $85.02 \pm 0.00$ & 6.00 \\
&Moderate$_B$ 
& $85.88 \pm 0.00$ & $85.49 \pm 0.00$ & $85.32 \pm 0.00$ & $84.48 \pm 0.00$ & 7.00 \\
&SCOSS ~~~~~(ours)        
& $91.21 \pm 0.11$ & $90.37 \pm 0.10$ & $89.32 \pm 0.04$ & $\mathbf{88.52 \pm 0.13}$ & 3.38 \\
&SCOSS$_B$ ~~(ours)       
& $91.18 \pm 0.05$ & $90.40 \pm 0.12$ & $89.36 \pm 0.11$ & $88.48 \pm 0.22$ & 2.88 \\
&SCOSS$_{CC}$ (ours)     
& $91.21 \pm 0.17$ & $90.31 \pm 0.10$ & $\mathbf{89.42 \pm 0.23}$ & $88.45 \pm 0.04$ & 3.13 \\

\hline
\end{tabular}
}
\end{table}

\vspace{-3mm}

\section{Conclusions}

This work investigated coreset selection strategies for image classification using SGC and SVM classifiers. The results demonstrate that it is possible to substantially reduce the training set size and consequently training time and GPU memory. Using only 20\% of the original training data substantially reduced SGC training time by approximately 74\% and GPU memory consumption by 64\% on CIFAR-10. On CUB-200, this efficiency comes with an accuracy trade-off, decreasing from 63.57\% to 53.07\%. However, the computational savings of more than half provide an attractive trade-off for resource-constrained scenarios. The ensemble mechanism improves accuracy and robustness by reducing coreset selection variability, with larger gains when run-to-run variability is higher. On CUB-200, SGC benefits more from coreset selection at lower sampling ratios, likely due to the fine-grained nature of the dataset. On CIFAR-10, which has fewer and larger classes, class balancing plays a less prominent role, and the ensemble tends to reduce differences between methods. Among the evaluated methods, SCOSS$_B$ achieves the best or near-best performance for SGC in the majority of cases, while unbalanced Moderate Coreset is often the weakest baseline, suggesting that covering the full score distribution with class-balanced stratified sampling is an effective coreset selection strategy.

Several directions remain open for future work, including testing alternative scoring criteria (beyond Euclidean distance), evaluating additional baselines, extending experiments to other datasets and classifiers, and exploring alternative ensemble aggregation strategies. A limitation of this work is the computational cost of running the full experimental evaluation, which involves multiple runs and ensemble repetitions across several methods, variants, and sampling ratios.

\section*{Acknowledgments}
Pedro Rocha Dantas is supported by a fellowship from the University of São Paulo (USP, PRPI Ordinance No.~1032, ``\emph{Apoio aos Novos Docentes}''). This work was also financially supported by the São Paulo Research Foundation (FAPESP, grant \#2025/10602-5) and the Institute of Mathematics and Computer Science (ICMC-USP).

\section*{Origin of the Work}
This paper presents ongoing work from an undergraduate research project conducted by the student Pedro Rocha Dantas.

\section*{AI Usage Declaration}
The authors used AI-based language models (Claude, Gemini, and ChatGPT) exclusively for reviewing and improving the clarity of text written by the authors.

\bibliographystyle{sbc}
\bibliography{sbc-template}

\end{document}